\documentclass[letterpaper]{article}
\usepackage{spconf,amsmath,graphicx}
\usepackage{mwe}
\usepackage{textcomp}
\usepackage{siunitx}
\usepackage[toc,page]{appendix}

\usepackage[utf8]{inputenc} %
\usepackage[T1]{fontenc}    %
\usepackage{hyperref}       %
\usepackage{url}            %
\usepackage{booktabs}       %
\usepackage{amsfonts}       %
\usepackage{nicefrac}       %
\usepackage{microtype}      %
\usepackage{algorithm}
\usepackage{algorithmic}
\usepackage{graphicx}
\usepackage{subcaption}
\usepackage{xcolor}
\usepackage{bm}
\usepackage{mathtools}
\usepackage{adjustbox}
\usepackage{algorithm}
\usepackage{algorithmic}

\title{Enhanced Adversarial Strategically-Timed Attacks against Deep Reinforcement Learning}
\name{Chao-Han Huck Yang$^{1}$, Jun Qi$^{1}$, Pin-Yu Chen$^{2}$, Yi Ouyang$^{3}$,  I-Te Danny Hung$^{4}$, Chin-Hui Lee$^{1}$, Xiaoli Ma$^{1}$}
\address{1. Georgia Institute of Technology, Atlanta, GA, USA\\ 2. IBM Research, Yorktown, NY, USA \\ 3. Preferred Network America, Berkeley, CA, USA\\ 4. Columbia University, NY, USA}
\begin{document}
\maketitle
\begin{abstract}
Recent deep neural networks based techniques, especially those equipped with the ability of self-adaptation in the system level such as deep reinforcement learning (DRL), are shown to possess many advantages of optimizing robot learning systems (e.g., autonomous navigation and continuous robot arm control.) However, the learning-based systems and the associated models may be threatened by the risks of intentionally adaptive (e.g., noisy sensor confusion) and adversarial perturbations from real-world scenarios. In this paper, we introduce timing-based adversarial strategies against a DRL-based navigation system by jamming in physical noise patterns on the selected time frames. To study the vulnerability of learning-based navigation systems, we propose two adversarial agent models: one refers to online learning; another one is based on evolutionary learning. Besides, three open-source robot learning and navigation control environments are employed to study the vulnerability under adversarial timing attacks. Our experimental results show that the adversarial timing attacks can lead to a significant performance drop, and also suggest the necessity of enhancing the robustness of robot learning systems.  
\end{abstract}
\section{Introduction}
\label{sec:intro}

Deep Reinforcement Learning (DRL) has gained a widespread applications in digital gaming, robotics and control. In particular, the main DRL approaches, such as the value-based deep Q-network (DQN) ~\cite{mnih2015human}, Asynchronous Advantage Actor-Critic (A3C)~\cite{mnih2016asynchronous}, and the population-based Go-explore~\cite{ecoffet2019go}, have succeeded in mastering many dynamically unknown action-searching environments~\cite{ecoffet2019go}. Relying on the similarity between the adaptive and interacting behaviors, DRL-based models are commonly used in the domain of navigation and robotics, and achieve a noticeable improvement over classical methods. However, despite the significant performance enhancement, DRL-based models may incur some new challenges in terms of system robustness against adversarial attacks. For example, the DRL-based navigation systems are likely to propagate and even enlarge risks (e.g., delay, noisy, and pixel-wise pulsed-signals~\cite{yang2019causal} on the sensor networks of vehicle~\cite{johansen2015estimation}) induced from the attackers. Besides, unlike the image classification tasks where only a single mission gets involved, the navigation learning agent has to deal with a couple of dynamic states (e.g., inputs from sensors or raw pixels) and the related rewards. Our work mainly focuses on the robustness analysis of strategically-timed attacks by potential noises incurred from the real world scenarios. More specifically, we formulate the adversarial attacks on two DRL-security settings:
\begin{itemize}
\item White-box attack: if attacker can access to model parameters, some potential function needs to be used to estimate the learning performance to jam in noise. 
\item Black-box attack: without the requirements of model parameters, the attacker trains a policy agent with the opposite reward objective via observing actions from the victim DRL network, the state, and the reward from the environment. 
\end{itemize}

\begin{figure}[ht!]
\begin{center}
   \includegraphics[width=1.0\linewidth]{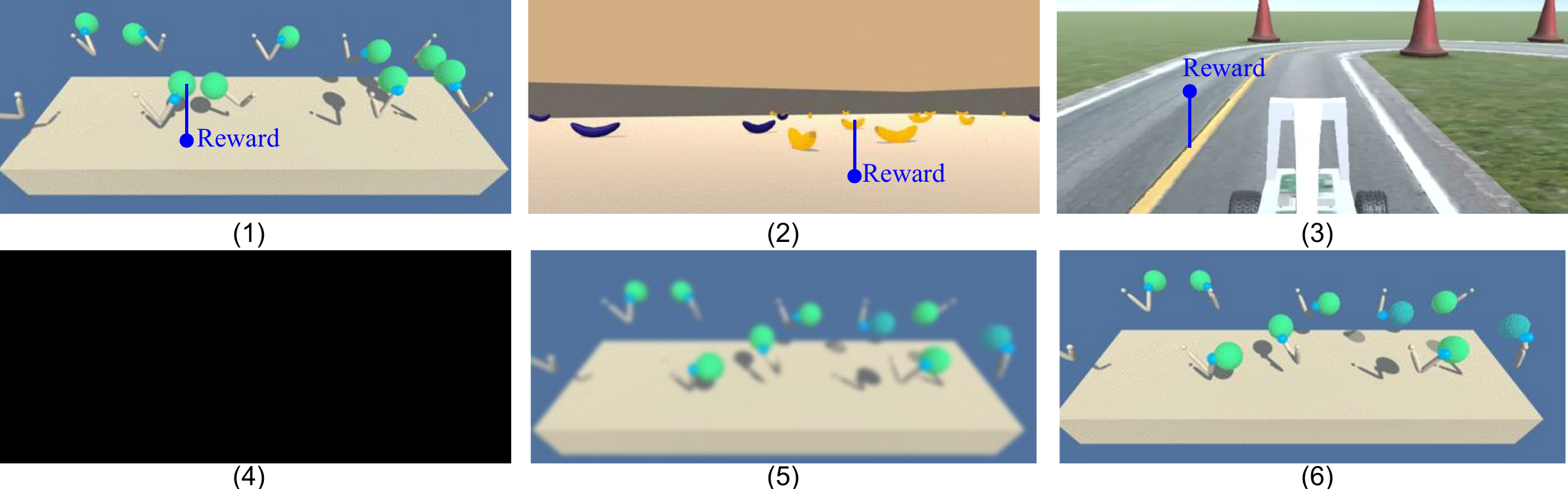}
\end{center}
\caption{The 3D robot learning environments: (1) continuous robot arm control as the $Env_1$; (2) banana collector as the $Env_2$; (3) self-driving donkey car as the $Env_3$. Noisy observation under timing attack: (4) zero-out; (5) random sensor fusion; (6) adversarial perturbation.} 
\label{fig:figure1}
\vspace{-0.5cm}
\end{figure}

To validate the adversarial robustness of a navigation system, we attempt a new and important research direction based on a $3$D environment of (1) continuous robot arm control (e.g., Unity Reacher); (2) sensor-input navigation system (e.g., Unity Banana Collector~\cite{juliani2018unity}); (3) raw images of self-driving environments (e.g., Donkey Car) as shown in Fig.\ref{fig:figure1} (a), (b), and (c).

\section{Related work}
\label{sec:pagestyle}

\textbf{Scheduling Physical Attacks on Sensor Fusion.}
Sensor networks for the navigation system are susceptible to flooding-based attacks like Pulsing Denial-of-Service (PDoS)~\cite{luo2005new} and adversary selective jamming attacks~\cite{proano2010selective}. The related work includes the security and robustness of background noise, spoofing pulses, and jamming signals on autonomous vehicles. For example, Yan et al.~\cite{yan2016can} show that PDoS attacks can feasibly conduct on a Tesla Model S automobile equipped with standard millimeter-wave radars, ultrasonic sensors, forward-looking cameras. Besides, to detect any anonymous network attacks, a sensing engine defined by some offline algorithms is required within a built-in network system. Furthermore, a recent work~\cite{cao2019adversarial} also demonstrates that the LiDAR-based Apollo-Auto system~\cite{fan2018baidu} could be fooled by adversarial noises during the 3D-point-cloud pre-processing phase as a malicious reconstruction.\\
\textbf{Adversarial Attacks on Deep Reinforcement Learning.}
Many works are denoted to adversarial attacks on neural network classifiers in either white-box settings or black-box ones~\cite{goodfellow2014explaining, chen2017zoo}. Goodfellow et al.~\cite{kurakin2018adversarial} proposed adversarial examples for evaluating the robustness of machine learning classifiers. Zeroth order optimization (ZOO)~\cite{chen2017zoo} was employed to estimate the gradients of black-box systems for generating adversarial examples. Besides, the tasks on RL-based adversarial attacks aim at addressing policy misconduct~\cite{huang2017adversarial,lin2017tactics} or generalization issues~\cite{pinto2017robust}. In particular, Lin et al.~\cite{lin2017tactics} developed a strategically-timed attacking method in which at time $t$, an agent takes action based on a policy derived from a Potential Energy Function~\cite{mohri2012foundations}.
However, these approaches do not consider the update of online weights associated with the size of the action space. In this work, we further improve the potential estimated model from \cite{lin2017tactics} by weighted-majority online learning, which owns a performance guarantee with a bound for $regret_T$ in Eq. (\ref{eq:bound}). Besides, we introduce a more realistic black-box timed-attack setting.
\vspace{-4mm}
\label{sec:typestyle}

\section{Method}
\label{sec_attack}
\subsection{Noisy Observation from the Real World} We define \label{subsec_phy}
a noisy DRL framework of a robot learning system under perturbation, where a noisy state observation $noisy\_s_{t}$ can be formulated as the addition of a state $s_{t}$ and a noise pattern $noise(t)$: 
\begin{equation}
\label{eq:noise}
noisy\_s_{t} = s_t + noise(t).
\end{equation}
We propose three principal types of noise test (from $T_1$, $T_2$ to $T_3$ ) from the real world to impose adversarial timing attacks: \\
\textbf{~Pulsed Zero-out Attacks ($T_1$):} Off-the-shelf hardwares~\cite{yan2016can} can affect the entire sensor networks by an over-shooting noise $noisy\_s_{t} = 0$ incurred from a timing attack in Eq.(\ref{eq:noise}) as Fig. \ref{fig:figure1} (4).\\
\textbf{~Gaussian Average on Sensor Fusion ($T_2$):}
Sensor fusion is an essential part of the autonomous system by combining of sensory data from disparate sources with less uncertainty. We define a noisy sensor fusion system by a Gaussian filter for getting $noisy$\_$s_{t}$ in Eq.(\ref{eq:adv:rl}) and shown as Fig. \ref{fig:figure1} (5).\\
\textbf{~Adversarial Noise Patterns ($T_3$):} Inspired by the fast gradient sign method (FGSM) \cite{goodfellow2014explaining, huang2017adversarial} based DQN attacks, we use FGSM to generate adversarial patterns against the prediction loss of a well-trained DQN. We use $\epsilon = 0.3$ and a restriction of $\ell_{\infty}$-norm, where $x$ is the all input including $s_{t}$ and $r_{t}$; $y$ = $a_{t}$ is an optimal output action by weighting over possible actions in Eq.(\ref{eq:adv:rl}):
\begin{equation}
\label{eq:adv:rl}
noise(t) = \epsilon \operatorname{sign}\left(\nabla_{x} J(\theta,x,y)\right).
\end{equation}
\textbf{To evaluate the performance of each timing selection algorithm in following sections, each model will receive noise patterns (from $T_1$, $T_2$ to $T_3$) and average the total reward as Table \ref{tab:my-table}.} In a perspective of system level, we take the random pulsed-signal as a attacking baseline. We jam in PDoS signals discussed in Sec. \ref{subsec_phy} randomly with maximum constrains $\mathbf{H}$ times (we use $\mathbf{H}$ = $40$ from \cite{lin2017tactics} as a baseline) to block agent from obtaining actual state observations in an episode.
\subsection{Enhanced White-Box Strategically-Timed Attack by Online Learning}
\textbf{White-box adversarial setting.} Recently, since various pre-defined DRL architectures and models (e.g., Google Dopamine \cite{castro2018dopamine}) are released for public use and as a key to Business-to-Business (B2B) solution, an adversarial attacker is likely to access the open-source and design an efficient strategically-timed attack. 
 \\
\textbf{Weighted-Majority Potential Energy Function.}
 We first propose an advanced adversarial attack which is originated from online learning and based on the algorithm of weighted majority algorithm (WMA). The procedures of WMA are shown in~Eq. \ref{eq:pot} and Algorithm \ref{alg:wma}, where we introduce $d$ experts for weighting the revenues incurred by taking $d$ actions. The weights of experts are equally initialized to $1$ and then iteratively updated as the step (12) in the Algorithm \ref{alg:wma}. At each time $t$, steps ($7$) and ($8$) suggest that we obtain both $a_{t}^{\max}$ and $a_{t}^{\min}$ which correspond to the actions of maximum and minimum costs. The decision of attacking the states relies on the threshold value $c(s_{t},  \textbf{w}_{t}, a_{t}^{\max}, a_{t}^{\min})$. If $c(s_{t}, \textbf{w}_{t}, a_{t}^{\max}, a_{t}^{\min})$ is greater than a pre-specified constant threshold $\beta$, we intend to attack the states by adding pulses to make the user have random observations. The choices of $c(s_{t}, \textbf{w}_{t}, a_{t}^{\max}, a_{t}^{\min})$ are based on the difference of two potential energy functions (inspired by \cite{lin2017tactics} and \cite{huang2017adversarial}) defined as~(\ref{eq:pot})\footnote{For potential energy estimation on policy-based model (e.g., A3C), we use a weighted-majority average as $c\left(s_{t}, w_{t}\right)=\max _{a_{t}} w^{T}_{t}\pi\left(s_{t}, a_{t}\right)-\min _{a_{t}} w^{T}_{t}\pi\left(s_{t}, a_{t}\right)$.}:

\begin{equation}
    \label{eq:pot}
    \begin{split}
c(s_{t}, \textbf{w}_{t}, a_{t}^{\min}, a_{t}^{\max}) &= \frac{\textbf{w}_{t}^{T}\exp(-\textbf{Q}(s_{t}, a_{t}^{\max}))}{\sum_{a_{t}^{(k)}}\textbf{w}_{t}^{T}\exp(-\textbf{Q}(s_{t}, a_{t}^{(k)}))} \\
&- \frac{\textbf{w}_{t}^{T}\exp(-\textbf{Q}(s_{t}, a_{t}^{\min}))}{\sum_{a_{t}^{(k)}}\textbf{w}_{t}^{T}\exp(-\textbf{Q}(s_{t}, a_{t}^{(k)}))} 
   \end{split}
\end{equation}
We use the strategically-timed attacks in~\cite{lin2017tactics} as a baseline with $\beta$ = 0.3 to evaluate our WMA-enhance algorithms. Then, we further discuss a learning bound for this advanced WMA-policy estimation. \\
\textbf{Proposition 1:} Assuming that the total number of rounds $T>2\log(d)$, the weighted algorithm enjoys the bound as Eq.(\ref{eq:bound}), where $Z_{t}$ denotes a normalization term at time $t$. 
\begin{equation}
\label{eq:bound}
\begin{split}
    &\hspace{4mm} regret_{T}\\
    &=  \sum\limits_{t=1}^{T}\frac{\textbf{w}_{t}^{T}\exp(-\textbf{Q}(s_{t}, a_{t}))}{Z_{t}} - \min\limits_{i\in [d]}\sum\limits_{t=1}^{T}\frac{\exp(-\textbf{Q}(s_{t}, a_{t}^{(i)}))}{Z_{t}} \\
    &\le \sqrt{2\log(d)T}.
    \end{split}
\end{equation}
Proposition $1$ \cite{mohri2012foundations} suggests that the weighted revenues are more likely to reach the global optimal in theory, since the regret at time $T$ is upper bounded by a constant value in Alg.\ref{alg:wma}.

\begin{algorithm*}[tb]
   \caption{Adversarial Online Learning Attack based on Weighted Majority Algorithm}
   \label{alg:wma}
\begin{algorithmic}
  \STATE {\bfseries 1.}  \textbf{Input}: number of experts, $d$; number of rounds, $T$, a threshold constant $\beta$. \\
  \STATE {\bfseries 2.} \textbf{Parameter}: $\eta = \sqrt{2\log (d) / T}$, expert weight $\textbf{w}\in R^{d}$ associated with $d$ actions.  \\
  \STATE {\bfseries 3.} \textbf{Initialize}: $\bar{\textbf{w}} = (1, 1, ..., 1)$ .\\
  \STATE {\bfseries 4.} \textbf{For} $t = 1, 2, ..., T$ \\
  \STATE {\bfseries 5.} \hspace{4mm} Set $\textbf{w}_{t} = \bar{\textbf{w}}_{t} / Z_{t}$, where $Z_{t} = \sum_{i} \bar{w}^{(i)}_{t}$. \\
  \STATE {\bfseries 6.} \hspace{4mm} Receive revenues from all experts $\textbf{Q}(s_{t}, a_{t}) = [Q(s_{t}, a_{t}^{(1)}), Q(s_{t}, a_{t}^{(2)}), ..., Q(s_{t}, a_{t}^{(d)})]$. \\
  \STATE {\bfseries 7.} \hspace{4mm} $a_{t}^{\min} \leftarrow \arg\min_{a_{t}} \textbf{w}_{t}^{T} \exp(-\textbf{Q}(s_{t}, a_{t}))$. \\
  \STATE {\bfseries 8.}\hspace{5mm}  $a_{t}^{\max} \leftarrow \arg\max_{a_{t}} \textbf{w}_{t}^{T} \exp(-\textbf{Q}(s_{t}, a_{t}))$.   \\ 
  \STATE {\bfseries 9.} \hspace{4mm} Compute the threshold function $c(s_{t}, \textbf{w}_{t}, a_{t}^{\max}, a_{t}^{\min})$. 
  \STATE {\bfseries 10.} \hspace{3mm} If $c(s_{t}, \textbf{w}_{t}, a_{t}^{\max}, a_{t}^{\min}) > \beta$:
  \STATE {\bfseries 11.} \hspace{8mm} Attack the state by $s_{t} \leftarrow $shuffle$(s_{t}) $ \\
  \STATE {\bfseries 12.} \hspace{3mm} \textbf{Update rule $\forall i$}, $\bar{\textbf{w}}_{t+1}^{(i)} = \bar{\textbf{w}}_{t}^{(i)}\exp(-\eta \textbf{Q}(s_{t}, a_{t}^{(i)}))$. \\
\end{algorithmic}
\end{algorithm*}

\begin{table*}[ht!]
\centering
\caption{A comparison of Performance of the timing attack algorithms on $Env_{1}$, $Env_{2}$, and $Env_{3}$ environments. We try to evaluate the robustness of a robot learning system with four types of strategically-timed attack algorithms, namely random selection; weighted majority algorithm (WMA); parameter exploring policy gradients~\cite{sehnke2010parameter} adversarial strategy agent (PEPG-ASA), and Lin et al.~\cite{lin2017tactics}. All experiments are tested for ten times under three different types of noise patterns (zero-out, Gaussian, and adversarial noises), where the total rewards are averaged by dividing $30$.}
\label{tab:my-table}
\begin{tabular}{|l|l|lll|l|}
\hline
\textbf{Model} & Baseline & Random & WMA & PEPG-ASA & Lin et al. \cite{lin2017tactics} \\ \hline
$Env_{1}$: Continuous Robot-Arm Control with \textbf{DQN}~\cite{mnih2015human} & 30.2$\pm$2.1 & 22.8$\pm$0.4 & \textbf{4.2}$\pm$\textbf{1.0} & 6.4$\pm$1.3 & 5.2$\pm$1.2 \\
$Env_{1}$: Continuous Robot-Arm Control with \textbf{A3C}~\cite{mnih2016asynchronous} & 30.1$\pm$3.6 & 23.2$\pm$0.5 & \textbf{3.2$\pm$0.7} & 5.2$\pm$1.0 & 5.6$\pm$1.3 \\ \hline
$Env_{2}$: 3D-Banana Collector Navigation with \textbf{DQN} & 12.1$\pm$2.1 & 10.8$\pm$2.8 & \textbf{3.2$\pm$2.3} & 7.4$\pm$1.9 & 6.9$\pm$1.6 \\
$Env_{2}$: 3D-Banana Collector Navigation with \textbf{A3C} & 12.1$\pm$1.6 & 9.6$\pm$1.7 & \textbf{3.4$\pm$1.1} & 5.3$\pm$1.4 & 5.2$\pm$1.3 \\ \hline
$Env_{3}$: Donkey Car Navigation with \textbf{DQN} & 1.2$\pm$0.1 & 0.8$\pm$0.5 & \textbf{0.2$\pm$0.1} & 0.4$\pm$0.2 & 0.4$\pm$0.1 \\
$Env_{3}$: Donkey Car Navigation with \textbf{A3C} & 1.1$\pm$0.4 & 0.8$\pm$0.2 & \textbf{0.3$\pm$0.2} & 0.6$\pm$0.3 & 0.6$\pm$0.2 \\ \hline
\end{tabular}
\end{table*}
\subsection{Black-Box Strategically-Timed Attack by Adversarial Evolutionary Strategy}
\textbf{Black-box adversarial setting.} Since an adversarial insidious attacking agent is hardly recognizable, an adversarial agent is able to drive the equilibrium of DRL-based system with an opposite objective reward without any information of targeted DRL-model. Thus, we propose an adversarial-strategic agent (ASA) via a population-based training method based on parameter exploring policy gradients~\cite{sehnke2010parameter} (PEPG) to optimize a black-box system. The PEPG-ASA algorithm can dynamically select sensitive time frames for jamming in an physical noise patterns in Section~\ref{subsec_phy}, which is likely to minimize the total system-rewards from an off-online observation of the input-output pairs without accessing actual parameters from the given DRL framework as below:
\begin{itemize}
\item observation: records of state $S$ from [$s_0$, $s_1$,..., $s_n$] and adversarial reward against victim navigation DRL-agent $R_{advs}$ from [$r_0$, $r_1$,..., $r_n$], an adversarial reward $R_{adv}$ as a black-box security setting.
\item adversarial reward $R_{adv}$: a negative absolute value of the environmental reward $R_{env}$. 
\end{itemize}
An obvious way to maximize $E[R_{adv}|s,a_{adv},\pi_{adv}] $ is to estimate $\nabla E$. Differentiating this form of the expected return with respect to $\rho$ and applying sampling methods, where $\rho$ in Eq. (\ref{eq:eq3:pepg}) are the parameters determining the distribution over $\theta$, the agent can generate h from $p(h|\theta)$ and yield the following gradient estimator:
\begin{equation}
\nabla_{\boldsymbol{\rho}} E(\boldsymbol{\rho}) \approx \frac{1}{N} \sum_{n=1}^{N} \nabla_{\boldsymbol{\rho}} \log p(\boldsymbol{\theta} | \boldsymbol{\rho}) r\left(h^{n}\right).
\label{eq:eq3:pepg}
\end{equation}
The probabilistic policy, which is parametrized over a single parameter $\theta$ for PEPG, has the advantage of taking deterministic actions such that an entire track of history can be traced by sampling the parameter $\theta$.
\section{Results}
\label{others}
\subsection{3D Control and Robot Learning Environment Setup}
Our testing platforms were based on the most recently released open-source `Unity-3D' environments \cite{juliani2018unity} for robotic applications.\\
\textbf{$\mathbf{Env_1}$~Reacher:} A double-jointed arm could move to the desired position. A reward of +0.1 is provided for each step that the agent's hand is in the goal location. The observation space consists of 33 variables corresponding to the position, rotation, velocity, and angular velocities of the arm. Every action is a vector with four numbers, corresponding to torque applicable to two joints. Each entry in the action vector should be a numerical value between -1 and 1.\\
\textbf{$\mathbf{Env_2}$~Banana Collector:} A reward of $+1$ is provided for collecting a yellow banana, and a reward of $-1$ is provided for collecting a blue banana from a first-person view vehicle to collect as many yellow bananas as possible while avoiding blue bananas. The state-space has 37 dimensions and contains the agent's velocity, along with the ray-based perception of objects around the agent's forward direction. Four discrete actions are available to associate with four moving directions.\\ 
\textbf{$\mathbf{Env_3}$~Donkey Car:} Donkey Car is an open-source embedded system for radio control vehicles with an off-line RL simulator. The state input is the image from the front camera with 80 $\times$ 80 pixels, the actions are equal to two steering values ranging from -1 to 1, and the reward is a cross-track error (CTE). We use a modified reward from ~\cite{prakash2019use} divided by 1k to balance track-staying and maximize its speed. 
\subsection{Performance Evaluation}
\begin{figure}[ht!]
\begin{center}
   \includegraphics[width=1.0\linewidth]{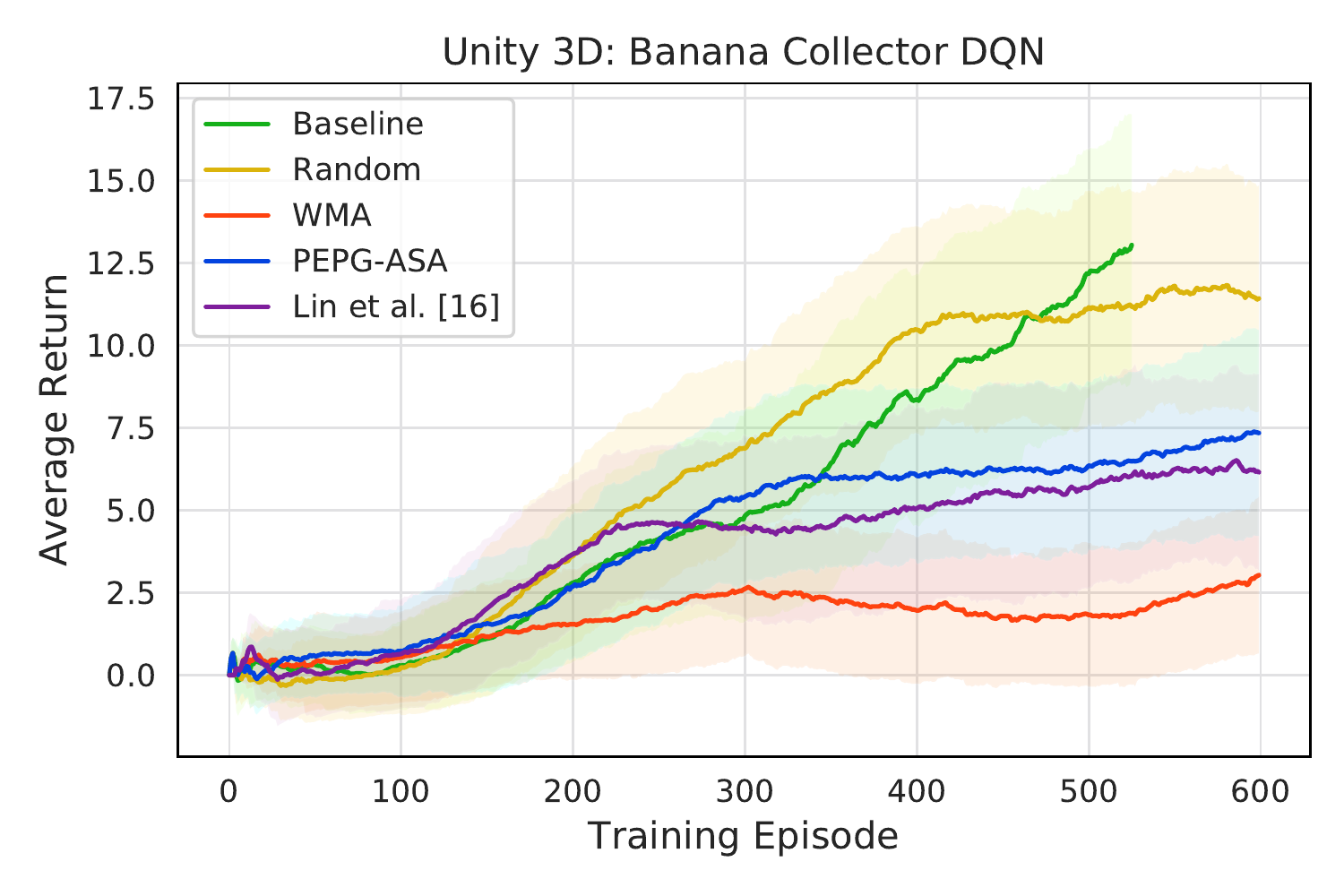}
\end{center}
\caption{Learning performance of a DQN agent testing in Unity 3D banana collector, a 3D-navigation task, included the baseline (a.k.a. no attacks); random jamming, WMA, PEPG-ASA, and the method from Lin et. al. } 
\label{fig:figure3}
\vspace{-0.5cm}
\end{figure}

We applied two classical DRL algorithms, namely DQN and A3C, to evaluate the learning performance relative to well-trained DRL models in Tab. \ref{tab:my-table}.\\
\textbf{Baseline (aka no attack):}
We modify DQN and A3C models from the open-source Dopamine 2.0~\cite{castro2018dopamine} package to avoid an overparameterized model with reproducibility guarantee.\\ 
\textbf{Adversarial Robustness (aka under attack):}
Assuming the presence of one adversarial attacker, we highlight some important results. Overall, although the WMA (white-box setting) outperforms the PEPG-ASA (black-box setting), it also requires much more information of a navigation system during the online potential-energy estimation and training.
In Fig. \ref{fig:figure3}, we show a result of DQN evaluate on the four types of attack method compared with the baseline performance, a random noise injector (Random), WMA, PEPG-ASA, and Lin~\cite{lin2017tactics} shown in Tab. \ref{tab:my-table}. WMA shows a stable threaten result as a competitive attack method.
\\

\vspace{-4mm}
\label{sec:typestyle}

\section{Conclusion}
This work introduces two novel adversarial timing attacking algorithms for evaluating DRL-based model robustness under white-box and black-box adversarial settings. The experiments suggest that the improved performance of DRL-based continuous control and robot learning models can be significantly degraded in adversarial settings. In particular, both valued and policy-based DRL algorithms are easily manipulated by a black-box adversarial attacking agent. Besides, our work points out the importance of the robustness and adversarial training against adversarial examples in DRL-based navigation systems. Our future work will discuss the visualization and interpretability of robot learning and control systems in order to secure the system. To improve model defense, we could also adapt the adversarial training \cite{goodfellow2014explaining} to train DQN \& A3C models by noisy states.  
\vfill\pagebreak

\clearpage
\bibliographystyle{IEEEbib}
\bibliography{strings,refs}

\end{document}